\pdfoutput=1

\documentclass[11pt]{article}

\usepackage[final]{acl}

\usepackage{times}
\usepackage{latexsym}

\usepackage[T1]{fontenc}

\usepackage[utf8]{inputenc}

\usepackage{microtype}

\usepackage{inconsolata}

\usepackage{graphicx}

\usepackage{comment}
\usepackage{titling}
\usepackage{multirow}
\usepackage{array}
\usepackage{wrapfig}
\usepackage{amsmath}
\usepackage{xcolor}
\usepackage{tcolorbox}
\usepackage{makecell}

\usepackage{blindtext}
\newcommand\blfootnote[1]{%
\begingroup
\renewcommand\thefootnote{}\footnote{#1}%
\addtocounter{footnote}{-1}%
\endgroup
}

\usepackage{color, colortbl}
\definecolor{Yellow}{rgb}{1.0,0.98,0.8}

\title{Demonstrations Are All You Need:\\
Advancing Offensive Content Paraphrasing using In-Context Learning}

\author{
    \begin{tabular}{c c c}
        Anirudh Som & Karan Sikka & Helen Gent \\
        \normalfont\href{mailto:anirudh.som@sri.com}{anirudh.som@sri.com} & \normalfont\href{mailto:karan.sikka@sri.com}{karan.sikka@sri.com} & \normalfont\href{mailto:helen.gent@sri.com}{helen.gent@sri.com} \\ \\
        Ajay Divakaran & Andreas Kathol & Dimitra Vergyri \\
        \normalfont\href{mailto:ajay.divakaran@sri.com}{ajay.divakaran@sri.com} & \normalfont\href{mailto:andreas.kathol@sri.com}{andreas.kathol@sri.com} & \normalfont\href{mailto:dimitra.vergyri@sri.com}{dimitra.vergyri@sri.com} \\ \\
        & SRI &
    \end{tabular}
}

\begin{document}
\maketitle


\begin{abstract}

Paraphrasing of offensive content is a better alternative to content removal and helps improve civility in a communication environment. Supervised paraphrasers; however, rely heavily on large quantities of labelled data to help preserve meaning and intent. They also often retain a large portion of the offensiveness of the original content, which raises questions on their overall usability. In this paper we aim to assist practitioners in developing usable paraphrasers by exploring In-Context Learning (ICL) with large language models (LLMs), \emph{i.e.}, using a limited number of input-label demonstration pairs to guide the model in generating desired outputs for specific queries. Our study focuses on key factors such as -- number and order of demonstrations, exclusion of prompt instruction, and reduction in measured toxicity. We perform principled evaluation on three datasets, including our proposed Context-Aware Polite Paraphrase (CAPP) dataset, comprising of dialogue-style rude utterances, polite paraphrases, and additional dialogue context. We evaluate our approach using four closed source and one open source LLM. Our results reveal that ICL is comparable to supervised methods in generation quality, while being qualitatively better by 25\% on human evaluation and attaining lower toxicity by 76\%. Also, ICL-based paraphrasers only show a slight reduction in performance even with just 10\% training data.  

\end{abstract}
\section{Introduction}
\emph{Disclaimer: Figures and examples in this work may feature offensive language.}

\blfootnote{Approved for public release: Distribution unlimited.}
\begin{figure}[t!]
    \centering
    \includegraphics[width=0.99\linewidth]{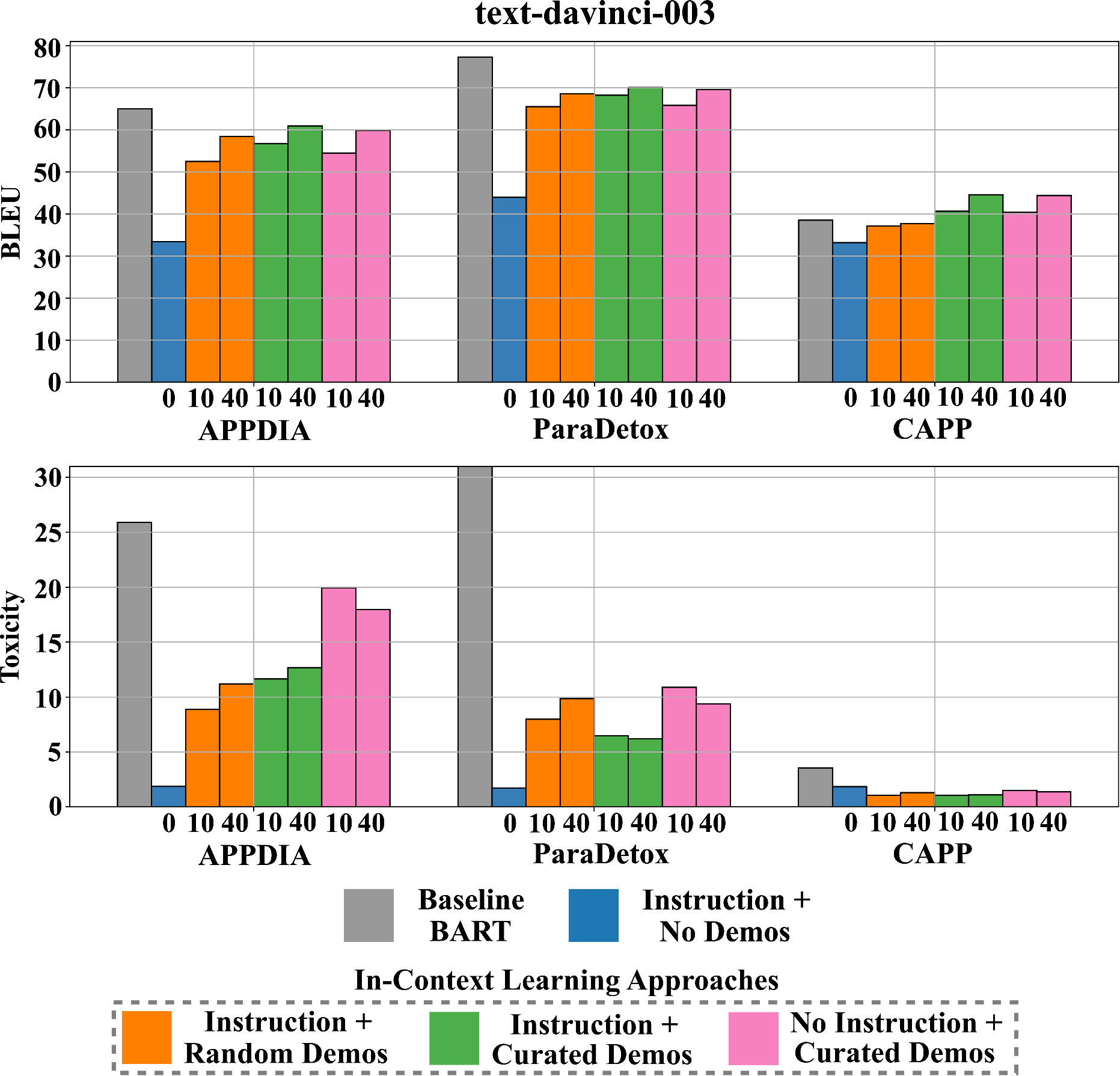}
    \caption{Influence of number and order of demonstrations, and instruction, on BLEU Score performance and measured Toxicity using the text-davinci-003 model. Comparison is done between BART, instruction-only prompting, and three In-Context Learning approaches. Numbers on the $x$-axis represent number of demonstrations used in the In-Context Learning framework. Note, measured Toxicity for BART in ParaDetox is 82, exceeding the set $y$-axis limit.}\label{fig_zero-random-curated}
    \vspace{-0.1in}
\end{figure}

Timely moderation helps curb the spread of hateful content on social-media platforms and prevents the harmful effects it has on a user's psychological well-being \cite{waldron2012harm, ye2023multilingual}. Unfortunately, the sheer volume of content generated on these platforms makes it infeasible to enforce a scalable human moderation process \cite{hassan2022studying, dosono2019moderation}. AI-based moderation systems can help with this problem. However, current systems often remove or flag offensive content, which can reduce user participation and diversity in online discussions \cite{xiang2012detecting, warner2012detecting, kwok2013locate, wang2014cursing, burnap2015cyber, nobata2016abusive, davidson2017automated, founta2019unified, jhaver2019did, ye2023multilingual}. A better alternative is to paraphrase offensive content to make it less offensive. 
Paraphrasing offensive content; however, is nontrivial since the paraphrased output should not only be inoffensive but also retain the original meaning and intent. 

Prior works \cite{atwell2022appdia,logacheva2022paradetox} have proposed using supervised generative models \cite{vaswani2017attention} like BART \cite{lewis2019bart}, to paraphrase offensive content. However, these methods require sufficient labelled training data, which makes it harder to adapt them to novel settings. 
Moreover, these models are optimized to perform well on certain automated metrics \cite{papineni2002bleu, zhang2019bertscore, lin2004rouge, vedantam2015cider} at the expense of possibly retaining a portion of the original toxicity, thereby making us question their overall usability for the targeted task (see Figure \ref{fig_zero-random-curated}).


The emergence of few-shot \emph{In-Context Learning} (ICL) has revolutionized the field by complementing the generalization capabilities of \emph{Large Language Models} (LLMs) to quickly and accurately adapt to new tasks. It does this by using a small amount of labeled data, known as \emph{demonstrations} or \emph{demos} or \emph{examples} \cite{brown2020language}. As shown in Figure \ref{fig_zero-random-curated}, ICL approaches show BLEU score performance that is comparable to BART, but significantly reduces the measured toxicity \cite{Detoxify}. Through detailed, principled experiments we explore the viability of ICL for paraphrasing offensive content, which to the best of our knowledge has not been done before. Our key contributions and findings in this paper are summarized below.


\begin{enumerate}
    \itemsep0em 
    
    \item Influence of the following factors on generation quality, as briefly shown in Figure \ref{fig_zero-random-curated}. \newline(a) \emph{Number of Demonstrations:} Performance improves by increasing number of demos but eventually saturates.
    \newline(b) \emph{Selection and Order of Demonstrations:} Systematically selecting and ordering demos is better than its random counterpart. It is more effective to select demos that are semantically similar to the query and curate them in a decreasing/increasing order of similarity. 
    \newline(c) \emph{Exclusion of Prompt Instruction in Prompt:}  ICL without the main instruction only slightly affects performance but at the expense of toxicity. Thus we need both demonstrations and instructions to simultaneously preserve performance and lower toxicity. 
    \newline(d) \emph{Robustness to Training Data Size:} Carefully ordering demos shows robustness to available training data size, with only small decrease in generation performance even when 10\% of training data is only made available.
    
    \item We tested the capabilities of OpenAI's \emph{text-davinci-003, gpt-3.5-turbo-0613, gpt-3.5-turbo-instruct, gpt-3.5-turbo-1106} models and the open-source \emph{Vicuna-13b} model \cite{vicuna2023}. 
    ICL generated paraphrases are comparable to SOTA supervised approaches in performance, but on average show 76\% less toxicity and are 25\% better using a manual qualitative assessment, and thus have superior overall usability. We also show that our demonstration curation approach is simpler and faster than other more sophisticated methods that offer only marginal performance improvements at the expense of significant time delays.
    

    \item Current paraphrasers are less effective at mitigating offensiveness like rudeness in conversations. They are trained using datasets that focus on social-media content, and hence aren't directly applicable to dialogue-based environments. To this end we release a new \emph{Context-Aware Polite Paraphrase (CAPP)} dataset$^1$, a dialogue-style corpus of rude utterances and corresponding polite paraphrases, with samples accompanied by additional context in the form of prior turns from the dialogue. We conduct experiments to show the importance and benefit of incorporating context to improve paraphraser performance.
\end{enumerate}


\blfootnote{$^1$The CAPP dataset and generated paraphrases are available online at \href{https://github.com/anirudhsom/CAPP-Dataset}{https://github.com/anirudhsom/CAPP-Dataset}.}

\noindent\textbf{Paper Outline:} Section \ref{section-background} describes ICL in our experimental setting; details about selecting and ordering the demos; and finally our proposed CAPP dataset in detail. Section \ref{section-experiments} contains detailed experimental results. Section \ref{section-related-work} discusses related work. Section \ref{section-conclusion} concludes the paper.

\section{Method}\label{section-background}

\subsection{In-Context Learning}\label{subsection_icl-prompt-design}
Prompts used for ICL contain three parts -- (1) an instruction $I$ that defines the task to be performed; (2) a set of $n$ demonstrations from the training corpus, $D=(x_i,y_i)_{i=1}^{n}$, where $(x_i, y_i)$ denotes the offensive, inoffensive sentence pair; and (3) the offensive test query sample $x_q$. Consider the following prompt example with $n=2$ demonstrations, where the final sentence represents the query for which we want to generate the paraphrase.

\scalebox{0.75}{
\begin{tcolorbox}[width=3.5in]
\noindent\colorbox{pink}{\emph{\textbf{Instruction:}} Paraphrase the following sentence to}
\noindent\colorbox{pink}{be more polite.} \\
\noindent\colorbox{yellow}{\emph{\textbf{Sentence:}} What's wrong with you?}
\noindent\colorbox{yellow}{\emph{\textbf{Paraphrase:}} Are you feeling alright?}
\noindent\colorbox{yellow}{\emph{\textbf{Sentence:}} Get out of the way.}\\
\noindent\colorbox{yellow}{\emph{\textbf{Paraphrase:}} Can you please step aside?}
\noindent\colorbox{green}{\emph{\textbf{Sentence:}} What's the matter with you?}\\
\noindent\colorbox{green}{\emph{\textbf{Paraphrase:}}}
\end{tcolorbox}}

The impact of each part on the BLEU score and toxicity is briefly illustrated in Figure \ref{fig_zero-random-curated}. For instance, prompts with only instruction show the lowest BLEU scores, followed by prompts with only demos, while prompts that include both have the best BLEU scores. In terms of Toxicity, prompts with just instruction show the least Toxicity, followed by prompts that include both demos and instruction, while prompts that only include demos exhibit a higher toxicity. The order of demos is also crucial, and we discuss this next. 

\subsection{Selection and Ordering of Demonstrations}\label{subsection_semantic-similarity}

Here we describe our approach to select and order the demonstrations. We first compute normalized vector embeddings for each training sample $x_i$ and the query $x_q$, denoted as $e_i$ and $e_q$ respectively. Next, the cosine similarity score between $e_q$ and each $e_i$ is used to select $n$ demonstrations. We explored the following two variations for selecting the demonstrations -- (1) \emph{Least Similar}, (2) \emph{Most Similar}, \emph{i.e.}, select $n$ demos with the lowest and highest cosine similarity scores, respectively. These are compared to randomly selecting $n$ demos, that are arranged in no particular order. We further investigated if arranging the $n$ selected demos in either ascending or descending order based on their measured cosine similarity, had any impact on the overall performance. Using BLEU and toxicity, Figure \ref{fig_zero-random-curated} compares \emph{Random} selection to the \emph{Most Similar (Descending order)} approach, with the latter being better on both fronts. Our findings for other selection and ordering approaches are described in detail in Section \ref{subsection-order}.



\subsection{Context-Aware Polite Paraphrase (CAPP) Dataset}\label{section-data-collection}

Existing datasets \cite{atwell2022appdia, logacheva2022paradetox} contain comments flagged for toxicity and provide non-toxic paraphrases that maintain the core meaning in a neutral manner. However, they are not directly suitable to address rudeness in speech, as speech is often directed at specific participants, while social media posts have a broader audience, resulting in different styles and tones. Additionally, most social media posts can be remedied by removing explicit insults, but rude speech requires additional modifications to make it more polite. For instance, we should not just eliminate offensive language and direct insults in a rude utterance, but also transform an accusation of ignorance into an inquiry about knowledge. 

\begin{table}[t!]
    \centering
    \scalebox{0.9}{\begin{tabular}{|c||p{5.5cm}|}
    \hline
    \textbf{Score} & \hspace{0.65in}\textbf{Description} \\
    \hline
    \hline
    \multirow{2}{*}{5} & Perfect meaning-preserving polite paraphrase. \\
    \hline
    
    \multirow{2}{*}{4} & Paraphrase that is polite but somewhat distinct in meaning. \\
    \hline
    
    \multirow{2}{*}{3} & Meaning-preserving paraphrase that could be more polite. \\
    \hline
    
    \multirow{3}{*}{2} & Paraphrase that is very different in meaning and somewhat more polite than the original. \\
    \hline
    
    \multirow{3}{*}{1} & Paraphrase that is very different in meaning and not more polite than the original. \\
    \hline
    
    \end{tabular}}
    \vspace{-0.05in}
    \caption{Description of the scoring guidelines used for evaluating the CAPP dataset in Section \ref{section-data-collection}. The same guidelines were used again in Section \ref{subsection-quality} to evaluate quality of paraphrases generated by the different paraphrasers on the CAPP dataset.}\label{table-scoring_dataset}
    \vspace{-0.15in}
\end{table}

To address the aforementioned differences, we constructed a dialogue-style rude speech dataset by leveraging the OpenSubtitles corpus \cite{lison2016opensubtitles2016}. 
Our approach involved a three-step process to extract target rude utterances. First, we fine-tuned a DistilBERT-base model \cite{sanh2019distilbert} using both the Stanford Politeness corpus \cite{danescu2013computational} and a subset of manually labeled OpenSubtitles samples to train a three-class model capable of predicting polite, neutral, or rude sentences. Next, we use the fine-tuned model to annotate a larger, different portion of the OpenSubtitles corpus, bootstrapping additional training data for our final rudeness detection model. Finally, a separate portion of the OpenSubtitles dataset was selected and labeled as rude, polite, or neutral using the updated rudeness detection model, resulting in an intermediate set containing rude samples without polite paraphrases. Detailed information about the training/evaluation of the rudeness detector is provided in Appendix \ref{appendix_dataset-details}. When available, context in the form of prior turns from the dialogue that precede the rude utterance was also collected for the selected rude samples.

The gpt-3.5-turbo-0613 model was used to generate the Gold-Standard or Groundtruth polite paraphrases. To do this we first explored three different prompts for generating three versions of polite paraphrases before finally deciding on one -- (1) \emph{Context-Free:} No prior dialogue context was included in the prompt, ensuring that the generated paraphrase is solely based on the rude utterance; (2) \emph{Context-Infused:} Prompt includes context which can significantly influence the generated paraphrase; (3) \emph{Context-Aware:} Prompt includes context, with the generated paraphrase being less impacted by it. For each version, 500 rude utterances and their corresponding polite paraphrases were randomly selected for qualitative evaluation. An in-house annotator assessed the quality of the paraphrases using the scoring guidelines in Table \ref{table-scoring_dataset} and was not informed about the type of prompt used to generate the polite paraphrases. The annotator identifies as a 28 year old cis female (pronouns she/her) and was compensated monetarily. Table \ref{table-dataset-evaluation} shows the final evaluation scores. The Context-Aware prompt achieves a score comparable to the Context-Free prompt while still incorporating context like the Context-Infused prompt. Context-Aware combines the benefits of both, and was hence used in the CAPP dataset.

\begin{table}[t!]
    \centering
    \scalebox{0.9}{
    \begin{tabular}{|c||c|}
        \hline
        \textbf{Prompt} & \textbf{Manual Evaluation Score}$\uparrow$ \\
        \hline
        \hline
        Context-Free & $4.214\pm1.047$\\
        \hline
        Context-Infused & $3.324\pm0.839$\\
        \hline
        Context-Aware & $4.096\pm1.093$\\
        \hline
    \end{tabular}
    }
    \vspace{-0.05in}
    \caption{Human evaluation scores of 500 polite paraphrases generated using different prompts. A higher score indicates a qualitatively better approach.}\label{table-dataset-evaluation}
    \vspace{-0.15in}
\end{table}

\section{Experiments and Discussion}\label{section-experiments}

We realized ICL using OpenAI's text-davinci-003, gpt-3.5-turbo-0613 models and their latest stand-ins, and the open-source Vicuna-13b model. We performed evaluation on the APPDIA \cite{atwell2022appdia}, ParaDetox \cite{logacheva2022paradetox}, CAPP datasets, with the corresponding (\#training, \#test) samples being (1584, 199), (11927, 670), (7939, 1120) respectively. APPDIA contains offensive Reddit comments and their inoffensive paraphrases. The ParaDetox corpus consists of toxic and non-toxic sentence pairs, obtained by filtering the ParaNMT corpus \cite{wieting2017paranmt}. In CAPP, 55\% of the training set and 53\% of the test set contains prior dialogue context information. We used the sentence transformer (\emph{all-mpnet-base-v2}) \cite{reimers-2019-sentence-bert} to generate the normalized embeddings described in Section \ref{subsection_semantic-similarity}. We evaluated generation quality using
automated evaluation metrics such as BLEU \cite{papineni2002bleu}, BERT-F1 \cite{zhang2019bertscore}, ROUGE \cite{lin2004rouge} and CIDEr \cite{vedantam2015cider}. For Toxicity we used the implementation by \cite{Detoxify}. The exact prompt instruction used in all experiments is provided in Appendix \ref{appendix-instrcution-prompts}.

\begin{figure}[t!]
    \centering
    \includegraphics[width=0.99\linewidth]{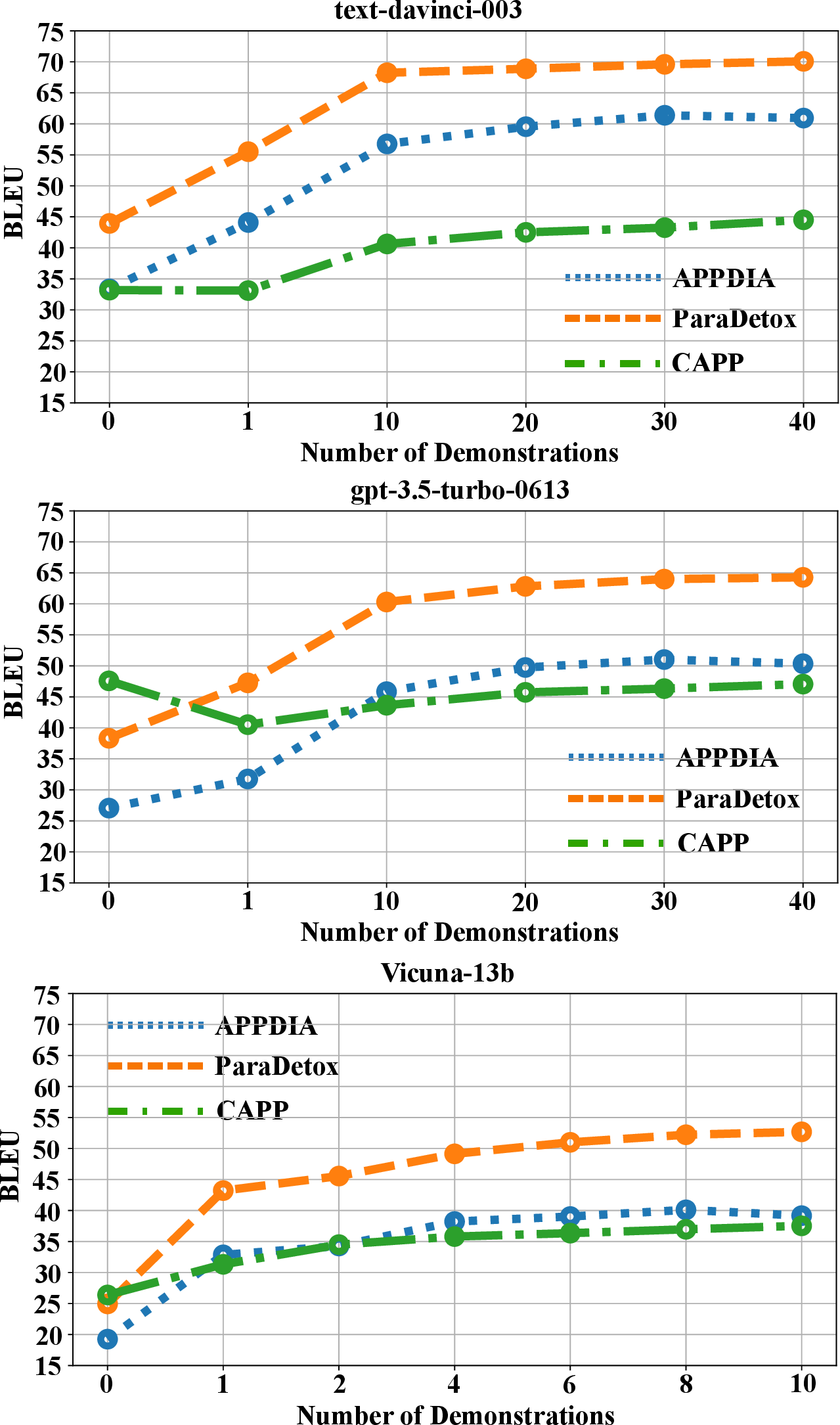}
    \caption{BLEU as a function of number of demos. Noticeable improvement in BLEU is observed in the beginning, with performance saturating after a certain number of demos.}
    \label{fig_number-of-examples}
    \vspace{-0.15in}
\end{figure}

\begin{figure*}[ht!]
    \centering
    \includegraphics[width=0.99\linewidth]{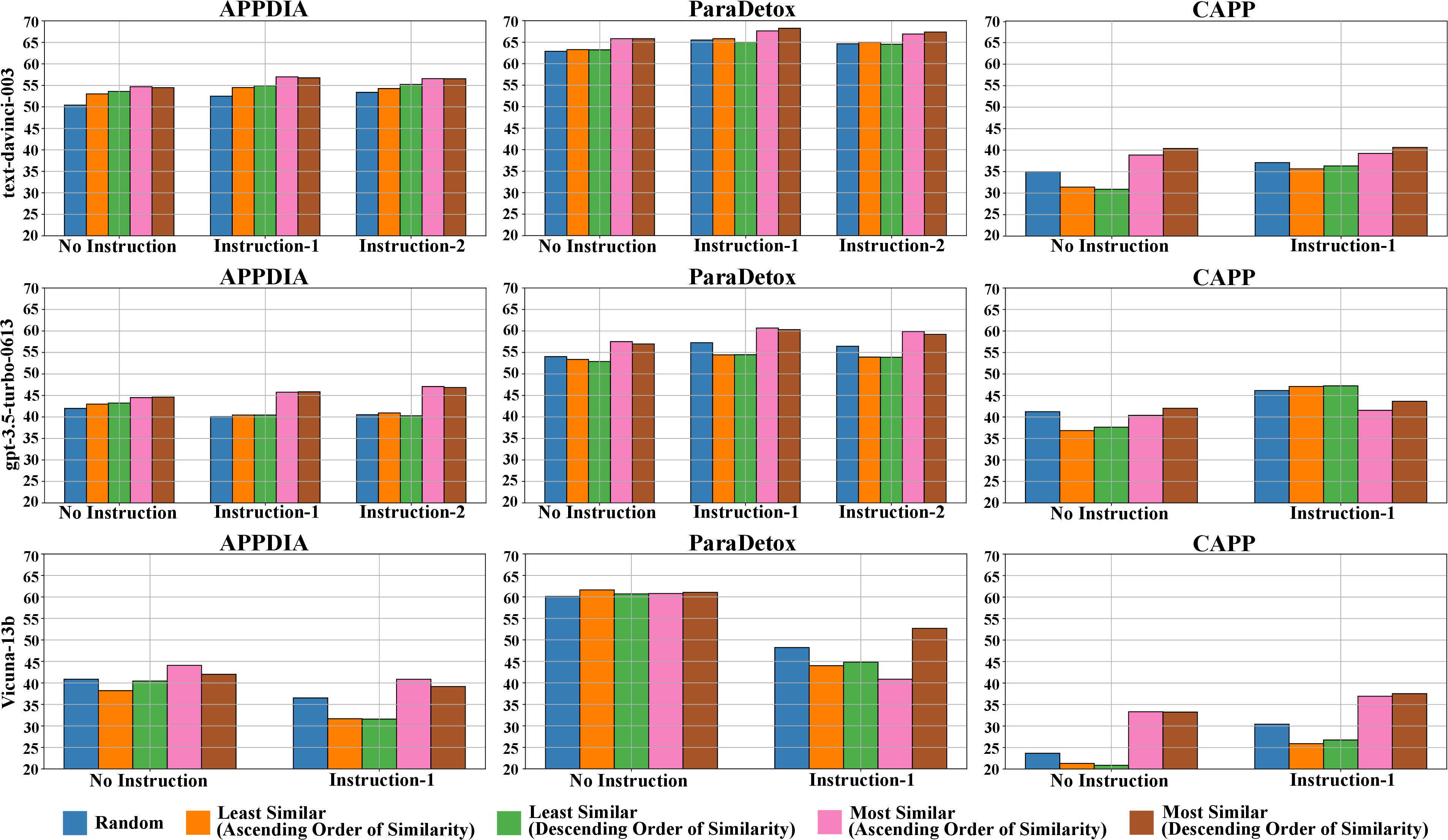}
    \caption{BLEU as a function of order of demonstrations and type of instruction used in the prompt design. Demonstrations that are semantically more similar to the query sample show better performance than less semantically similar and randomly selected samples. Also, prompts that only include demonstrations (\emph{i.e.,} \emph{No Instruction}) show a BLEU score that is comparable to prompts that include instruction and demonstrations.}
    \label{fig_order-of-examples}
    \vspace{-0.15in}
\end{figure*}
\subsection{Number of Demonstrations}\label{subsection-number}

Figure \ref{fig_number-of-examples} shows the relation between number of demonstrations and BLEU (refer to Appendix \ref{appendix-section-number}, Figure \ref{fig_appendix_number-of-examples} for other metrics). We set the number of demonstrations to $[0,1,10,20,30,40]$ for text-davinci-003, gpt-3.5-turbo-0613, and $[0,1,2,4,6,8,10]$ for Vicuna-13b.  
We used the proposed \emph{Most Similar (Descending Order)} approach to select and order the demos. 
We observe that BLEU improves rapidly until 10 demos for the OpenAI models and 4 demos for the Vicuna-13b model across all datasets. Further increasing the demos only results in slight improvement, as each additional demo is semantically less similar to the query and thereby less important than the demonstrations selected before \cite{liu2021makes}.

We notice in the case of the gpt-3.5-turbo-0613 model on CAPP dataset that BLEU without any demos is better than with 40 demos. 
It's possible that the main instruction used here was less effective in the ICL paradigm, and that a different instruction could have increased the BLEU score, as seen later in Section \ref{subsection-instruction-absence}. However, we believe this happens because the Gold-Standard for CAPP was also generated using gpt-3.5-turbo-0613. This hints at the possibility of ICL not necessarily improving paraphrasing performance of LLMs, which in turn were used to generate the dataset. We see similar observations in the following sections as well.

\subsection{Selection and Order of Demonstrations}\label{subsection-order}
\begin{figure}[t!]
    \centering
    \includegraphics[width=0.99\linewidth]{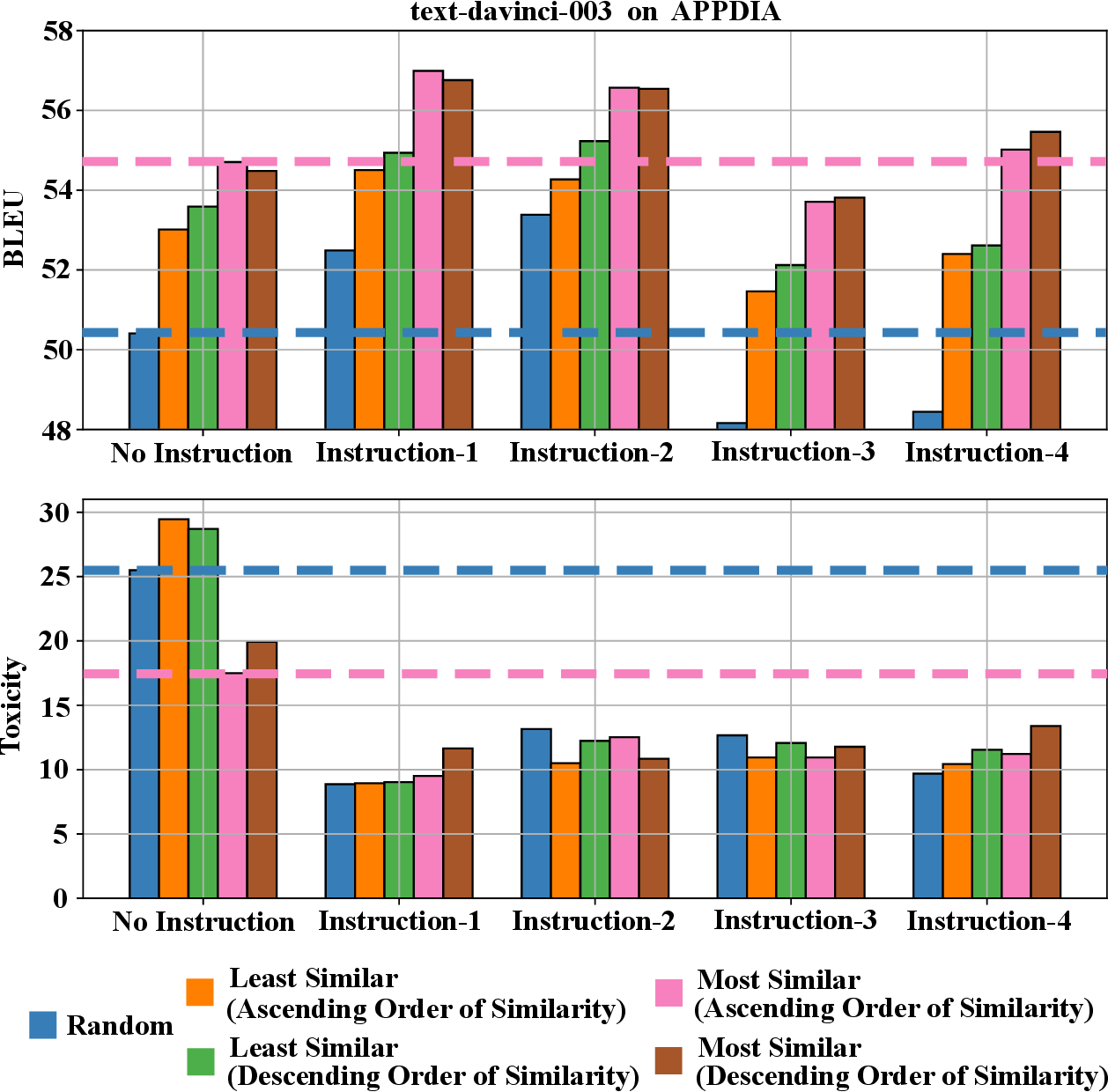}
    \caption{BLEU score and measured toxicity performance with different instructions but with the same set of demos. Instructions can either complement or work against the selected demos and accordingly affect the BLEU score. The \emph{No Instruction} setting shows comparable BLEU to prompts that include both instructions and demos but result in paraphrases with higher toxicity. The dotted reference lines are used to indicate the range in BLEU score under the \emph{No Instruction} setting.}
    \label{fig_bad-instruction}
    \vspace{-0.15in}
\end{figure}

We now discuss the effect of selection and ordering the demos in the prompt on BLEU. Note, in Figures \ref{fig_order-of-examples} and \ref{fig_bad-instruction}, the number of demonstrations was set to 10 and explore the different ordering mechanisms described in Section \ref{subsection_semantic-similarity}. In Figure \ref{fig_order-of-examples}, we observe that the \emph{Random} strategy sometimes achieves better BLEU than the \emph{Least Similar} strategy. While in most cases the \emph{Most Similar} shows better performance than both \emph{Random} and \emph{Least Similar}. This intuitively makes sense since \emph{Most Similar} represents samples from the training corpus that are most semantically similar to the query \cite{liu2021makes}. This enables the LLM to generate a paraphrase that is also similar to the Gold-Standard paraphrase of the query. Next, the order in which the demos are arranged also has an impact on BLEU score. We find that curating the demos in decreasing order of similarity often results in better BLEU than arranging them in increasing order of similarity. We also observe similar trends with other automated evaluation metrics. Note, the above observations do not apply to the gpt-3.5-turbo-0613 model on the CAPP dataset.

While the approaches described in Section \ref{subsection_semantic-similarity} are simple and effective, they might not bring out the best possible performance. Sophisticated methods to select and order demonstrations have been proposed and have shown better performance in other applications \cite{ye2022complementary,zhang2022active,lu2021fantastically}. However, we find that they offer only marginal improvement, while taking significantly longer times to process each query sample. For example, Table \ref{table-comparison} shows the different performance metrics and compute times for \cite{ye2022complementary}'s MMR method and the proposed ICL-based approach. Note, gpt-3.5-turbo-1106 was used as the generation model for MMR. While MMR might be marginally better quantitatively, it is several orders of magnitude slower than the proposed approach in terms of mean Similarity Compute Time and mean Demo Retrieval Compute Time. 
Please refer to Appendix \ref{appendix-section-order} for more details on this topic.

\begin{table*}[ht!]
\centering
\scalebox{0.605}
{\begin{tabular}{|c||c||c|c|c|c||c||c||c|c|}
    \hline
    \textbf{Dataset} & \textbf{Method} & \textbf{BLEU}$\uparrow$ & \textbf{BERT-F1}$\uparrow$ & \textbf{ROUGE}$\uparrow$ & \textbf{CIDEr}$\uparrow$ & \textbf{Toxicity}$\downarrow$ & \textbf{Quality}$\uparrow$ & \makecell{\textbf{Similarity}\\ \textbf{Compute Time}$\downarrow$} & \makecell{\textbf{Demo Retrieval}\\ \textbf{Compute Time}$\downarrow$}\\
    \hline
    \hline
    \parbox[t]{2mm}{\multirow{18}{*}{\rotatebox[origin=c]{90}{APPDIA}}} & \emph{Offensive Test Set} & - & - & - & - & 75.60 & - & \multirow{6}{*}{-} & \multirow{6}{*}{-} \\
    \cline{2-8}  & \emph{Inoffensive Gold-Standard} & - & - & - & - & 14.37 & 3.68$\pm$0.93  &  & \\
    \cline{2-8}& BART \cite{atwell2022appdia} & 65.0 & 68.1 & 65.6 & 4.77 & 25.91 & 3.42$\pm$1.08  &  & \\
    \cline{2-8}& T5 \cite{atwell2022appdia} & 65.3 & 69.2 & 66.5 & 4.75 & 20.15  & -  &  & \\
    \cline{2-8}& DialoGPT \cite{atwell2022appdia} & 42.3 & 46.7 & 38.0 & 1.11 & 14.51  & 3.52$\pm$0.93  &  & \\
    \cline{2-8}& PDTB+RST \cite{atwell2022appdia} & 46.2 & 50.7 & 42.5 & 1.54 & 16.39  & -  &  & \\
    \cline{2-10} & MMR-BERT (10 Demos) & 58.5 & 65.1 & 59.3 & 3.90 & 17.54 & - & 4.7538 & 0.0792 \\
    \cline{2-10}  & MMR-Embedding (10 Demos) & 57.9 & 63.8 & 57.7 & 3.79 & 14.11 & - & 0.0025 & 0.0786 \\

    \cline{2-10}&  gpt-3.5-turbo-1106 (10 Demos) &  57.1  &  64.0  &  57.20  &  3.72  &  \textbf{14.42}  & - & \multirow{10}{*}{\textbf{0.0025}} & \multirow{10}{*}{\textbf{0.0005}} \\
    \cline{2-8}&  gpt-3.5-turbo-1106 (40 Demos) & 61.2 & 66.4 & 61.9 & 4.23 & \textbf{19.55}  & - &  & \\
    \cline{2-8}& gpt-3.5-turbo-0613 (10 Demos) &  45.8  &  53.3  &  41.6  &  2.12  &  \textbf{7.00}  & \textbf{4.24$\pm$0.91} &  & \\
    \cline{2-8}& gpt-3.5-turbo-0613 (40 Demos) & 50.4 & 58.2 & 47.6 & 2.67 & \textbf{10.08}  & \textbf{4.11$\pm$1.00}  &  & \\
    \cline{2-8}&  gpt-3.5-turbo-instruct (10 Demos) & 51.9 & 58.9 & 50.5 & 2.81 & \textbf{15.86}  & - &  & \\
    \cline{2-8}& gpt-3.5-turbo-instruct (40 Demos) & 56.2 & 62.9 & 55.8 & 3.42 & \textbf{21.37}  & - &  & \\
    \cline{2-8}& text-davinci-003 (10 Demos) & 56.8 & 63.6 & 57.6 & 3.70 & \textbf{11.64}  & \textbf{3.98$\pm$1.05}  & & \\
    \cline{2-8}& text-davinci-003 (40 Demos) & 60.9 & 66.7 & 62.9 & 4.29 & \textbf{12.67}  & \textbf{3.77$\pm$1.08}  &  & \\
    \cline{2-8}& Vicuna-13b (4 Demos) &  38.2  &  46.8  &  34.9  &  1.41  &  \textbf{12.07}  & \textbf{3.87$\pm$1.00} &  & \\
    \cline{2-8}& Vicuna-13b (10 Demos) & 40.8 & 48.0 & 37.6 & 1.79 & \textbf{18.44}  & \textbf{3.91$\pm$1.07}  &  & \\
    \hline

    \hline
    \parbox[t]{2mm}{\multirow{15}{*}{\rotatebox[origin=c]{90}{ParaDetox}}}  & \emph{Offensive Test Set} & - & - & - & - & 88.64 & - & \multirow{3}{*}{-} & \multirow{3}{*}{-}\\
    \cline{2-8}  & \emph{Inoffensive Gold-Standard} & - & - & - & - & 6.56 & 3.77$\pm$0.97 &  & \\
    \cline{2-8}& BART \cite{logacheva2022paradetox} & 77.3 & 76.2 & 69.8 & 4.94 & 82.00 & 2.82$\pm$0.75  &  & \\
    \cline{2-10} & MMR-BERT (10 Demos) & 68.6 & 68.0 & 58.8 & 3.67 & 8.47 & - & 4.7538 &  0.5990 \\
    \cline{2-10}  & MMR-Embedding (10 Demos) & 67.6 & 67.3 & 57.7 & 3.52 & 8.91 & - & 0.0025 & 0.6010 \\
    \cline{2-10} &  gpt-3.5-turbo-1106 (10 Demos) & 67.6 & 67.3  & 57.2  &  3.45 &  \textbf{8.46}  & -  & \multirow{10}{*}{\textbf{0.0025}} & \multirow{10}{*}{\textbf{0.0048}}\\
    \cline{2-8} &  gpt-3.5-turbo-1106 (40 Demos) & 70.1 & 69.3 & 59.6 & 3.79 & \textbf{9.64}  & -  &  & \\
    \cline{2-8}& gpt-3.5-turbo-0613 (10 Demos) &  60.3  &  62.0  &  50.5  &  2.72  &  \textbf{5.71}  & \textbf{3.90$\pm$1.01} &  & \\
    \cline{2-8}& gpt-3.5-turbo-0613 (40 Demos) & 64.3 & 65.1 & 54.1 & 3.08 & \textbf{6.20}  & \textbf{3.92$\pm$1.02}  &  & \\
    \cline{2-8} &  gpt-3.5-turbo-instuct (10 Demos) & 65.2 & 66.4 & 55.6 & 3.17 & \textbf{9.97}  & - &  & \\
    \cline{2-8} & gpt-3.5-turbo-instuct (40 Demos) & 69.1 & 68.1 & 58.9 & 3.68 & \textbf{12.3}  & - &  & \\
    \cline{2-8}& text-davinci-003 (10 Demos) & 68.2 & 67.7 & 58.9 & 3.67 & \textbf{6.50}  & \textbf{4.34$\pm$0.91}  &  & \\
    \cline{2-8}& text-davinci-003 (40 Demos) & 70.1 & 69.3 & 60.4 & 3.95 & \textbf{6.21}  & \textbf{4.22$\pm$0.96}  &  & \\
    \cline{2-8}& Vicuna-13b (4 Demos) &  49.3  &  54.1  &  41.1  & 1.78   &  \textbf{7.23}  & \textbf{4.00$\pm$0.99}  &  & \\
    \cline{2-8}& Vicuna-13b (10 Demos) & 52.8 & 56.7 & 43.7 & 2.05 & \textbf{9.98}  & \textbf{4.53$\pm$0.84}  &  & \\
    \hline

    \hline
    \parbox[t]{2mm}{\multirow{16}{*}{\rotatebox[origin=c]{90}{CAPP}}}  & \emph{Offensive Test Set} & - & - & - & - & 25.87 & - & \multirow{4}{*}{-} & \multirow{4}{*}{-}\\
    \cline{2-8}  & \emph{Inoffensive Gold-Standard} & - & - & - & - & 0.94 & 4.38$\pm$0.83 &  & \\
    \cline{2-8}& BART & 38.5 & 48.3 & 36.3 & 1.86 & 3.54 & 3.78$\pm$0.87  &  & \\
    \cline{2-8}& T5 & 39.4 & 50.2 & 37.9 & 1.92 & 2.63  & 3.84$\pm$0.87  &  & \\
    \cline{2-10} & MMR-BERT (10 Demos) & 45.5 & 54.4 & 41.8 & 2.20 & 1.05 & - & 4.7538 &  0.3937 \\
    \cline{2-10}  & MMR-Embedding (10 Demos) & 44.0 & 52.4 & 39.9 & 2.10 & 1.34 & - & 0.0025 & 0.3936\\
    \cline{2-10} &  gpt-3.5-turbo-1106 (10 Demos) &  43.9  &  53.2  & 40.5   &  2.17  &  \textbf{1.22}  & - & \multirow{10}{*}{\textbf{0.0025}} & \multirow{10}{*}{\textbf{0.0031}}\\
    \cline{2-8} &  gpt-3.5-turbo-1106 (40 Demos) & 45.8 & 54.8 & 42.5 & 2.29 & \textbf{1.30}  & - & & \\
    \cline{2-8}& gpt-3.5-turbo-0613 (10 Demos) &  43.7  &  51.9  &  39.6  &  2.00  &  \textbf{0.82}  & \textbf{4.44$\pm$0.81} &  & \\
    \cline{2-8}& gpt-3.5-turbo-0613 (40 Demos) & 47.1 & 55.0 & 43.0 & 2.33 & \textbf{0.72}  & \textbf{4.58$\pm$0.76}  &  & \\
    \cline{2-8} & gpt-3.5-turbo-instruct (10 Demos) &  44.9  &  53.6  &  41.1  &  2.18  &  \textbf{1.23}  & - & & \\
    \cline{2-8} &  gpt-3.5-turbo-instruct (40 Demos) & 48.0 & 56.4 & 44.5 & 2.53 & \textbf{1.49}  & - & & \\
    \cline{2-8}& text-davinci-003 (10 Demos) & 40.6 & 49.6 & 36.1 & 1.73 & \textbf{1.04}  & \textbf{4.03$\pm$0.96}  &  & \\
    \cline{2-8}& text-davinci-003 (40 Demos) & 44.5 & 53.2 & 40.7 & 2.10 & \textbf{1.09} & \textbf{4.10$\pm$0.94}  &  & \\
    \cline{2-8}& Vicuna-13b (4 Demos) &  35.8  &  42.4  &  31.3  &  1.34  &  \textbf{1.04}  & \textbf{4.36$\pm$0.78} &  & \\
    \cline{2-8}& Vicuna-13b (10 Demos) & 37.5 & 35.9 & 33.6 & 1.55 & \textbf{1.02}  & \textbf{4.21$\pm$0.88}  &  & \\
    \hline
\end{tabular}}
\vspace{-0.05in}
\caption{Quantitative, qualitative and mean compute time (in seconds) assessment of different LLMs using the ICL paradigm and comparison against different baseline supervised approaches. Toxicity of the offensive test set and inoffensive ground-truth paraphrases is also provided. Differences in the reported Mean$\pm$Std Quality scores between each ICL-based approach and the different baselines is significantly different (\emph{i.e.,} $p\mathrm{-value}<0.05$).} 
\label{table-comparison}
\vspace{-0.15in}
\end{table*}

\begin{figure*}[t]
    \centering
    \includegraphics[width=0.99\linewidth]{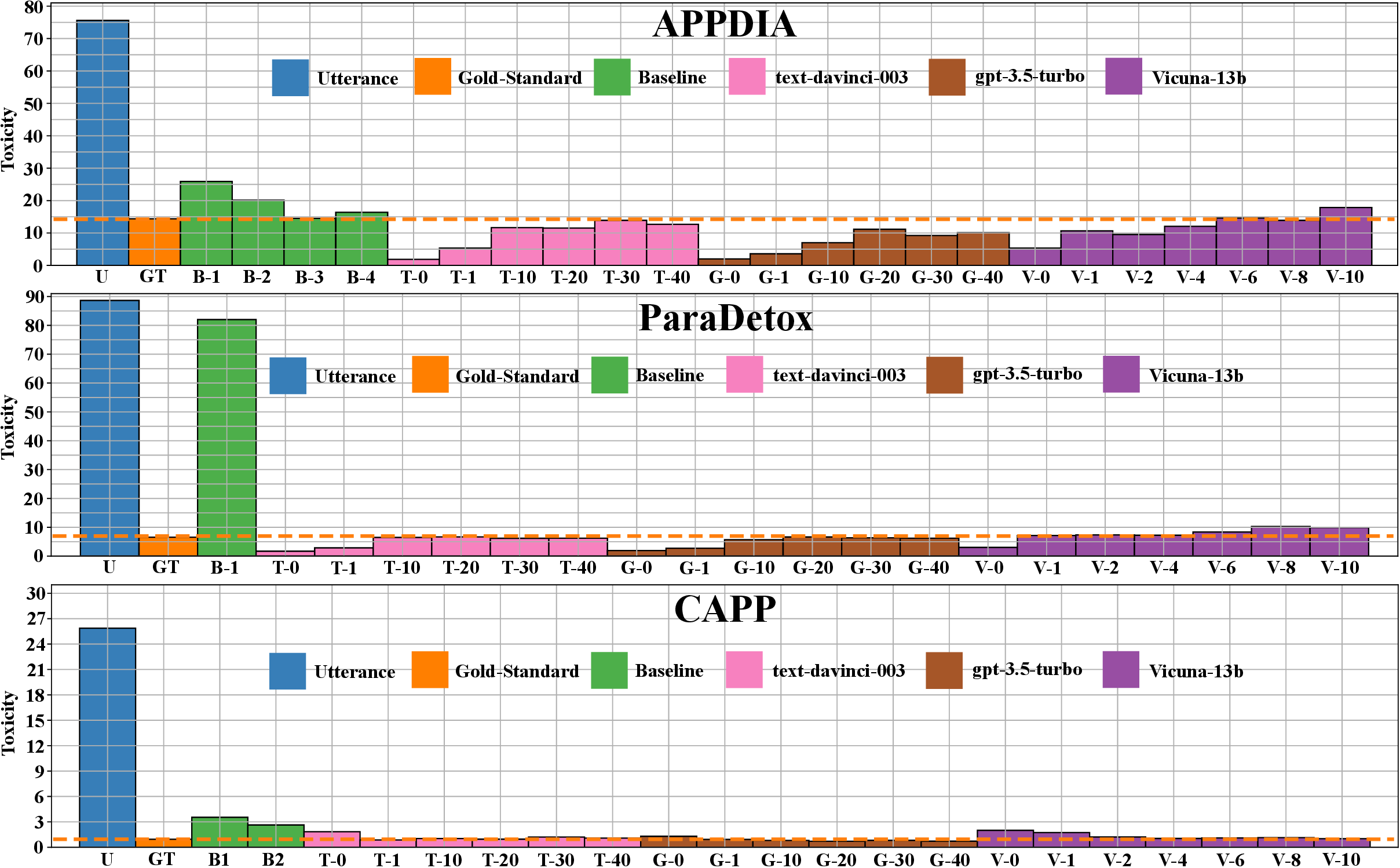}
    \caption{Average Toxicity measured using the \emph{Detoxify} \cite{Detoxify}. The orange dotted line serves as a reference for the Gold-Standard's Toxicity. U, GT, B-\#, T-\#, G-\#, V-\# along the $x$-axis refer to Utterance, Gold-Standard, Baseline methods, text-davinci-003, gpt-3.5-turbo, Vicuna-13b respectively. \# in T-\#, G-\#, V-\# indicate number of demonstrations used. Note, T-0, G-0, V-0 only contain an instruction in the prompt.}
    \label{fig_Toxicity}
    \vspace{-0.15in}
\end{figure*}

\subsection{Significance of Instruction}\label{subsection-instruction-absence}

We now investigate the effect of removing the instruction in the prompt. The left-most set of bars within each bar plot in Figures \ref{fig_order-of-examples} and \ref{fig_bad-instruction}, which show prompts with \emph{No Instruction}, display BLEU scores that are on par with prompts that include both instructions and demonstrations. 
This is interesting since it is a common practice to always include the instruction in the prompt even if no demonstrations are provided.
Our results suggest that when it is difficult to determine effective instructions for the target paraphrasing task, with ICL one can simply use a few systematically selected demonstrations to get high quality generated paraphrases.

In Figure \ref{fig_bad-instruction}, for text-davinci-003 on APPDIA, we observe that the \emph{No Instruction} setting retains a significant amount of the original content's toxicity, thereby making its usability questionable. Similar observations were made with other models (refer to Appendix \ref{appendix-section-instruction-absence}, Figure \ref{fig_appendix_bad_instruction}). Order of demos also plays an important role in the \emph{No Instruction} setting, with the \emph{Most Similar} showing much lower toxicity than both \emph{Random} and \emph{Least Similar} strategies. For cases that include both instruction and demos, the measured toxicity is less impacted by the order of demos, indicating that the main instruction serves as a toxicity regularizer.

We want to also highlight that creating a good instruction for paraphrasing tasks is non-trivial. Despite using good demos, a bad instruction can negatively impact the quality of the generated paraphrase. For instance, in Figure \ref{fig_order-of-examples}, the Vicuna-13b model shows better BLEU with just the curated demonstrations on the APPDIA and ParaDetox datasets. Similarly, in Figure \ref{fig_bad-instruction}, we see that certain instructions can result in lower BLEU than prompts that have \emph{No Instruction}.  

\subsection{Comparison with Supervised Approaches}\label{subsection-comparison}
We compare our ICL-based approach to prior state-of-the-art supervised baselines.
For APPDIA we use BART, T5, DialoGPT, and PDTB+RST methods as done in \cite{atwell2022appdia}; for ParaDetox we use BART as done in \cite{logacheva2022paradetox}; and for CAPP we fine-tuned BART-base and T5-base on the training set.  
  We used the default hyperparameters defined in the Transformers Seq2SeqTrainer for fine-tuning on CAPP.      
The comparison between our ICL-based approaches (including the newer OpenAI models, gpt-3.5-turbo-instruct and gpt-3.5-turbo-1106) and prior baselines is shown in Table \ref{table-comparison}. 

Note, the objective of any paraphraser should be to score high on generation quality and have a low Toxicity in the generated paraphrases. For APPDIA and ParaDetox, the BART and T5 models perform better than the ICL-based approach on the different standard evaluation metrics. However, the paraphrases generated by these baselines seem to retain a significant amount of the original toxicity. 
To better understand this issue, we use the Toxicity for the \emph{Inoffensive Gold-Standard} in each dataset as a point of reference. 
Ideally, a paraphraser should generate paraphrases whose average Toxicity is no greater than this reference. We observe that all baseline methods except DialoGPT show a higher Toxicity, while the ICL-based methods exhibit Toxicity that is lower or on par with that of the Gold-Standard. Our approach offers a better trade-off between generation quality and Toxicity.

Figure \ref{fig_Toxicity} illustrates the Toxicity measured for the different ICL-based methods by varying the number of demonstrations. The \emph{Most Similar (Descending Order)} strategy was used to select and organize the demos. It also displays the measured Toxicity of the Offensive Test Set, the Gold-Standard and the different baselines. Note that, B1, B2, B3, B4 refer to BART, T5, DialoGPT and PDTB+RST models respectively. For APPDIA and ParaDetox we observe that LLMs without any demonstrations show much lower Toxicity than when any demonstration is used. The reverse, however, is observed for the CAPP dataset. 
The absence of demos causes LLMs to fallback on their own task definition which results in paraphrases with Toxicity significantly different from that of the Gold-Standard. However, the absence of demos also causes the generated paraphrases to exhibit a lower BLEU score as seen earlier in Figure \ref{fig_number-of-examples}. A balance between the main instruction and demos can ensure generation of paraphrases that reduce offensiveness and score high using different automated metrics. 


\subsection{Additional Dialogue Context Helps}\label{subsection-add-context}

\begin{figure}[t!]
    \centering
    \includegraphics[width=0.99\linewidth]{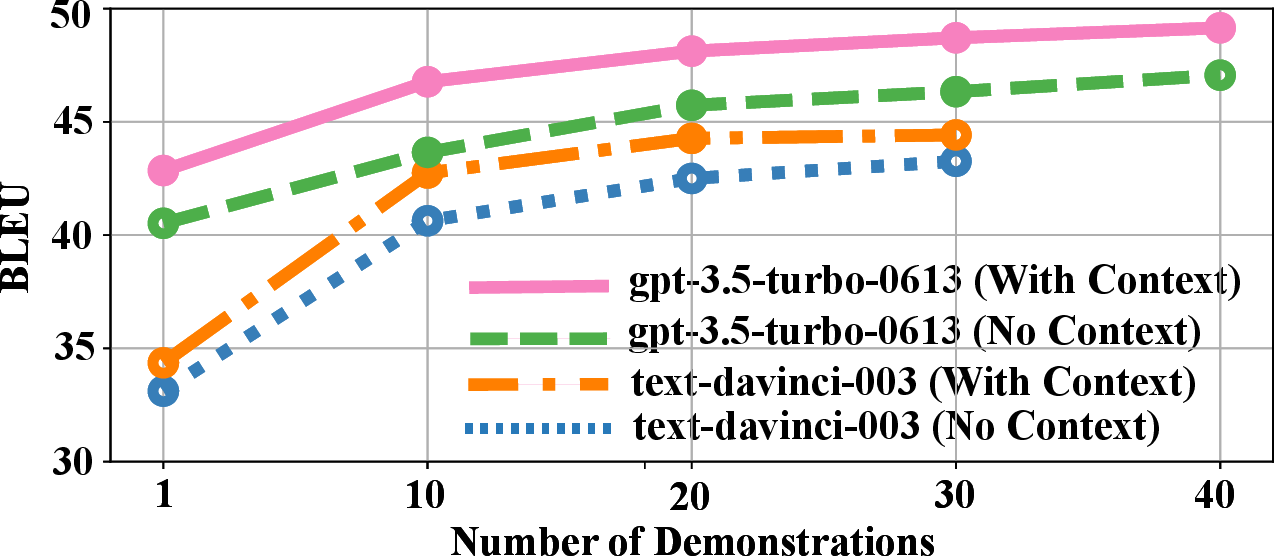}
    \caption{Comparison of BLEU performance on CAPP between including and excluding prior context in the form of prior dialogue utterances.}
    \label{fig_context-helps}
    \vspace{-0.in}
\end{figure}

We show preliminary results of using the prior utterances leading up to the rude utterance as additional context in the our ICL-based method. Similar to the example in Section \ref{subsection_icl-prompt-design}, we prepend the context for both the demo and the query. Figure \ref{fig_context-helps} shows BLEU score as we add/remove context and vary the number of demonstrations. We clearly see performance improvement by incorporating dialogue context using the text-davinci-003 and gpt-3.5-turbo-0613 models. We were unable to create a prompt for Vicuna-13b that successfully uses additional context in the ICL framework. We will develop such prompts for Vicuna-13b in future work. 

\subsection{Robustness to Reduced Training Data}\label{subsection-reduced-training-data}
\begin{figure}[t!]
    \centering
    \includegraphics[width=0.99\linewidth]{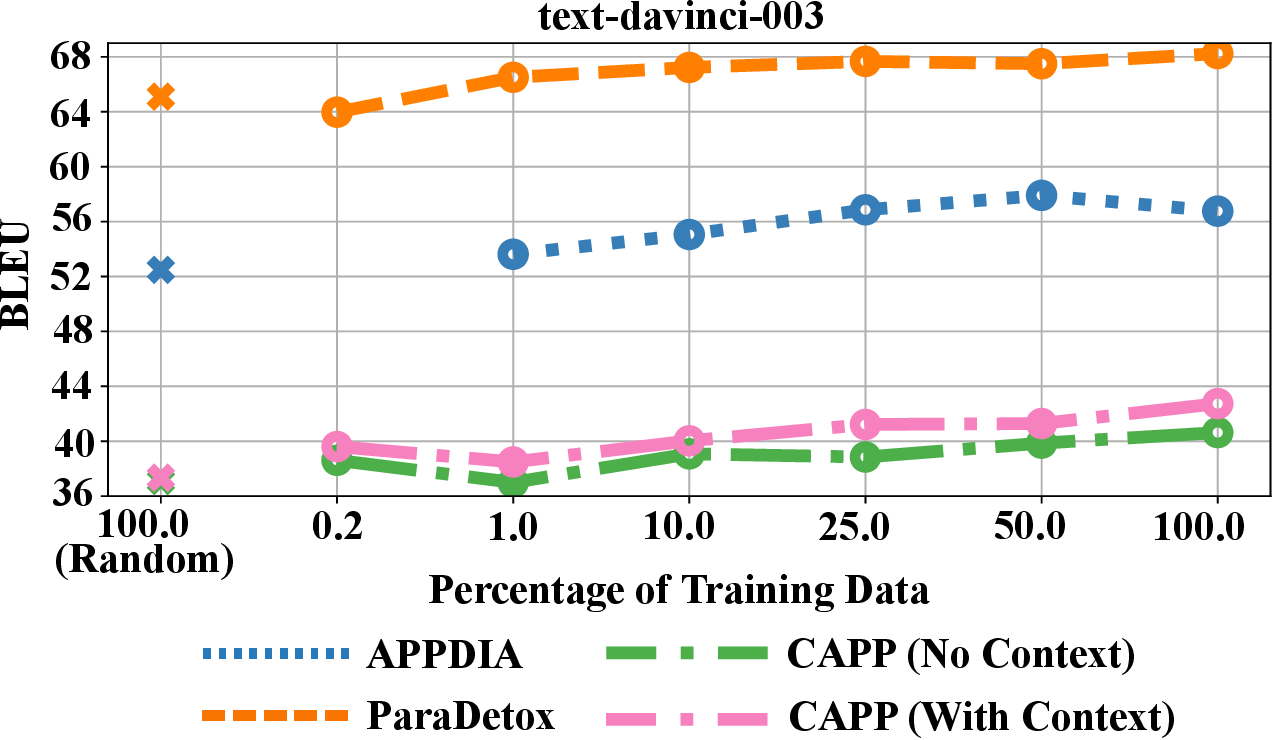}
    \caption{BLEU for the \emph{``Most Similar (Descending Order)''} approach as a function of percentage of training data available, and comparison with Randomly selected demonstrations using 100\% of the training data.}
    \label{fig_training-percentage}
    \vspace{-0.1in}
\end{figure}

Here we study the impact of available training data on the performance of our best performing strategy i.e.  \emph{Most Similar (Descending Order)}.
We observe only a minimal fall in BLEU up to 10\% of the training data as shown for text-davinci-003 in Figure \ref{fig_training-percentage} (refer to Appendix \ref{appendix-subsection-reduced-training-data}, Figure \ref{fig_appendix_training-percentage} for other models).
Further reducing training data results in noticeable drop in BLEU. We also find that reducing training dataset below 10\% results in BLEU score that is similar to \emph{Random} demo selection and arrangement strategy but with access to 100\% of the training data. That result shows that
our ICL-based method can work with limited training data and thus can be adapted quickly to novel settings.

\subsection{Manual Qualitative Assessment}\label{subsection-quality}

We also perform quality assessment of the Gold-standard and generated paraphrases using a human annotator. We select a subset of 150, 200, and 200 samples from the test-set of 
APPDIA, ParaDetox and CAPP respectively. 
For ParaDetox and CAPP, we use all the supervised baseline methods listed in Table \ref{table-comparison}. For APPDIA we only use BART and DialoGPT models for comparison. Our in-house annotator (mentioned earlier in Section \ref{section-data-collection}) used the scoring guidelines described in Tables \ref{table-scoring_dataset} and \ref{table-scoring_paraphrase}. Information about the type of model used to generate each paraphrase was not made available to the annotator. Table \ref{table-comparison} shows that the three ICL-based LLM models received a higher average score that is significantly different (\emph{i.e.,} $p$-value $<0.05$) than the corresponding baseline methods. 
We also note that Vicuna-13b's qualitative score was comparable to and in some cases better than the OpenAI models, despite having scored lower on metrics that are traditionally used to measure generation quality. 
This shows that open-source LLM paraphrasers are comparable to closed-source LLMs as per human assessment and might be a viable alternative. Refer to Appendix \ref{appendix-subsection-correlation} for additional analysis between manual evaluation score and toxicity metric.

\begin{table}[t!]
    \centering
    \scalebox{0.9}{\begin{tabular}{|c||p{5.5cm}|}
    \hline
    \textbf{Score} & \hspace{0.65in}\textbf{Description} \\
    \hline
    \hline
    \multirow{2}{*}{5} & Perfect meaning-preserving inoffensive paraphrase. \\
    \hline
    
    \multirow{2}{*}{4} & Paraphrase that is inoffensive but somewhat distinct in meaning. \\
    \hline
    
    \multirow{2}{*}{3} & Meaning-preserving paraphrase that could be less offensive. \\
    \hline
    
    \multirow{3}{*}{2} & Paraphrase that is very different in meaning and somewhat less offensive than the original. \\
    \hline
    
    \multirow{3}{*}{1} & Paraphrase that is very different in meaning and not less offensive than the original. \\
    \hline
    
    \end{tabular}}
    \vspace{-0.in}
    \caption{Description of the scoring guidelines used in Section \ref{subsection-quality} for evaluating the generated paraphrases for the APPDIA \cite{atwell2022appdia} and ParaDetox \cite{logacheva2022paradetox} datasets.}\label{table-scoring_paraphrase}
    \vspace{-0.1in}
\end{table}

\section{Related Work}\label{section-related-work}
Our paper explores the potential use of LLMs with ICL for paraphrasing systems. There has been  significant interest in better understanding the capabilities of ICL, but for other applications \cite{min2021noisy, zhao2021calibrate, razeghi2022impact, xie2021explanation, lampinen2022can, mishra2021reframing, chen2021meta, min2021metaicl, chen2023mixture}.  
For instance, \cite{lu2021fantastically} showed that order of demos has a significant impact on model performance. \cite{liu2021makes} showed that retrieving demonstrations that are semantically similar to the query can be a more effective approach to control the variability in performance. \cite{rubin2021learning} learned an encoding scheme to retrieve better demos for ICL. 
Other works also explored the influence of number of demos in different settings \cite{garg2022can, min2022rethinking, wei2023larger}. \cite{zhou2022large} evaluated the importance of each part in the prompt has towards the final performance. In this paper we study the impact of various components on the final performance, while ensuring that the toxicity of the outputs is within tolerable levels. This enables us to propose a few-shot solution to offensive content paraphrasing.
Most prior works \cite{atwell2022appdia, logacheva2022paradetox} have modeled paraphrasing as a sequence-to-sequence problem and trained models such as T5, BART on human annotated  data. Despite good generation results, these models tend towards higher toxicity and are difficult to adapt to new applications without collecting more data. Our solution addresses those challenges successfully, with only a fraction of the original training set.

\section{Conclusion}\label{section-conclusion}

In this paper, we focus on developing usable offensive content paraphrasing systems by leveraging generalization capabilities of LLMs and quickly adapting them to new tasks using ICL. A paraphraser should generate qualitatively good paraphrases that preserve the original content's meaning, while also minimizing toxicity. Focusing only on one of these aspects compromises overall usability. Compared to supervised approaches that require lot of training data and often produce undesired yet coherent paraphrases, our ICL-based framework is generally comparable on various evaluation metrics like BLEU, but is qualitatively better and helps significantly reduce toxicity in the generated paraphrases.  Through systematic experiments we tested the capabilities and limitations of ICL-based offensive paraphrasers. Other key highlights of using our ICL framework include: (1) Selection and arrangement of demos significantly impacts quality of paraphrases; (2) Measured toxicity is lowest when only the instruction is used and highest when only demos are used. Combining both instruction and demos helps ensure quality and usability of generated paraphrases; (3) Robust to limited data, \emph{i.e.,} with just 10\% training data we only see a slight decrease in overall performance, thereby enabling us to easily scale and deploy.

\newpage
\section*{Acknowledgments}

This material is based upon work supported by the Defense Advanced Research Projects Agency (DARPA) under Contract No. HR001122C0032. Any opinions, findings and conclusions or recommendations expressed in this material are those of the author(s) and do not necessarily reflect the views or policies of DARPA, the Department of Defense or the U.S. Government.

\section*{Limitations}

Here, we list the limitations identified in this paper:

\begin{enumerate}
    \item We found that ICL fails on datasets that were prepared using the same LLMs used in the ICL framework. Since we used the gpt-3.5-turbo-0613 model to create polite paraphrases for our CAPP dataset, we were unable to see the same observations in our results when using other models and datasets. This could have been avoided by creating manually annotated polite paraphrases. However, manual annotation is a laborious process and isn't scalable. Hence, a manual qualitative assessment was done on a small subset of the final CAPP dataset to ensure usability of the generated paraphrases.

    \item Prompt engineering for the Vicuna-13b model with the ICL framework is nontrivial. We found it difficult to create main instructions in the prompt that result in the Vicuna-13b model to behave in a desired way. Also, unlike the two OpenAI models, the number of demonstrations that can be effectively passed into Vicuna-13b is quite limited. In some cases we were able to concatenate more than 10 demos to the prompt but it often resulted in generating incomprehensible or empty outputs.

    \item The \emph{No Instruction} prompt explored in the paper resulted in paraphrases that are comparable to prompts that include both instruction and demos, on several automated evaluation metrics. However, we notice that the \emph{No Instruction} setting also retains a significant amount of toxicity from the original content. We propose that in situations where it is difficult to decide on a good main instruction, one could simply use a few carefully curated and ordered demos like the ``Most Similar (Descending Order)'' approach to generate paraphrases and check if it is within the desired toxicity levels.

    \item Our experimental results indicate that there is no single prompt that works in all situations. One must carefully balance the main instruction and the set of demos from the training corpus to get desired paraphrase outputs.

    \item We showed preliminary results showcasing the benefit of incorporating additional context in the form of prior two utterances in the ICL framework. We believe there can be better ways to incorporate this contextual information and further improve performance of LLMs.

    \item The closed-source OpenAI models are more powerful, faster and expensive to use. Despite open-source models like Vicuna-13b coming close to OpenAI models on other tasks, they still have a long way to go for offensive content paraphrasing.
\end{enumerate}

\section*{Ethics Statement}
We have to take great care with our collection of offensive content to protect privacy. We have to ensure judicious use of the collected data to protect the vulnerable against such speech. We recognize that our models cannot entirely eliminate offensive content from a given text. Additionally, we acknowledge that utilizing pretrained models may introduce biases in specific situations, as studies have revealed that pretrained models can be influenced by biases present in the data used for their initial training. We have to continue research on making sure that the LLMs do not hallucinate and end up injecting toxicity since we don't know what they have been trained on. There is a danger of this kind of technology being used in reverse, \emph{i.e.}, take harmless content and paraphrase to inject toxicity. We realize that ethics is an ongoing challenge. We are engaged with the Fairness, Accountability and Transparency community and are learning to address key ethics issues on an ongoing basis.

\newpage
\bibliography{custom}

\newpage
\appendix

\section{Context-Aware Polite Paraphrase (CAPP) Dataset}\label{appendix_dataset-details}

This section provides additional details about the training protocol used to develop the rudeness detector described in Section \ref{section-data-collection}. The initial rudeness detection model was trained on a randomly-selected subset of the OpenSubtitles corpus \cite{lison2016opensubtitles2016} that was manually annotated by an in-house human annotator, as well as the Stanford Politeness corpus \cite{danescu2013computational}. Between these two data sources, this model was trained on 10997 gold-quality samples. This intermediate model achieved an overall accuracy score of 59.40 across the three classes (rude, polite, and neutral). The accuracy for specifically rude utterances in the test data was 71.84 – higher than that of other classes. The final rudeness detection model, trained on a combination of the aforementioned gold data, and a larger subset (250379 samples) of the OpenSubtitles corpus annotated using the intermediate model, achieved an overall accuracy of 57.20 across the three classes, with an accuracy score of 71.84 on rude samples – still higher than the accuracy on other classes. The disparity in rudeness detector performance between classes was deemed acceptable due to its high performance on the rude class. Accurate identification of rude utterances was prioritized in the interest of generating a sufficient number and variety of rude utterances for paraphraser training – with an acceptance of the potential for noise in the form of some non-rude utterances in the training data.

\section{Instructions Used in Experiments}\label{appendix-instrcution-prompts}

The exact prompt instruction used for each experiment in Section \ref{section-experiments} is shown below. Except Subsection \ref{subsection-order} and \ref{subsection-instruction-absence}, we use only \textbf{Instruction-1} for the following  Subsections \ref{subsection-number}, \ref{subsection-comparison}, \ref{subsection-add-context}, \ref{subsection-reduced-training-data} and \ref{subsection-quality}.

\begin{enumerate}

    \item Instruction prompts used for text-davinci-003 and gpt-3.5-turbo models on the APPDIA dataset:
    \begin{enumerate}
        \item \textbf{Instruction-1:} Instruction: Rewrite the following sentence to be inoffensive
        \item \textbf{Instruction-2:} Instruction: Rewrite the following sentence to be inoffensive and ensure that the paraphrased sentence is similar to the original sentence
        \item \textbf{Instruction-3:} Instruction: Generate a paraphrased sentence that expresses a more balanced viewpoint
        \item \textbf{Instruction-4:} Instruction: Rewrite the following sentence with a more objective tone
    \end{enumerate}

    \item Instruction prompts used for text-davinci-003 and gpt-3.5-turbo models on the ParaDetox dataset: 
    \begin{enumerate}
        \item \textbf{Instruction-1:} Instruction: Rewrite the following sentence to be inoffensive
        \item  \textbf{Instruction-2:} Instruction: Rewrite the following sentence to be not toxic
    \end{enumerate}

    \item Instruction prompts used for text-davinci-003, gpt-3.5-turbo and Vicuna-13b models on the CAPP dataset: 
    \begin{enumerate}
        \item \textbf{Instruction-1:} The following sentence can contain rude text. Please provide one paraphrased sentence that is semantically similar to it and is polite.
    \end{enumerate}
    
    \item Instruction prompts used for Vicuna-13b model on APPDIA and ParaDetox datasets: 
    \begin{enumerate}
        \item \textbf{Instruction-1:} The following sentence contains offensive text. Please provide one paraphrased sentence that is semantically similar to it and is inoffensive.
    \end{enumerate}

    \item While incorporating additional context in the form of previous two turns in the dialogue, the instruction prompt used for text-davinci-003 and gpt-3.5-turbo models on the CAPP dataset: 
    \begin{enumerate}
        \item \textbf{Instruction-1:} Paraphrase only the below Sentence to be polite and semantically similar to the Sentence. Use the context as as reference but do not include any part of it in the final paraphrase.
    \end{enumerate}

\end{enumerate}

\section{Experiments}\label{appendix-experiments}

\subsection{Number of Demonstrations}\label{appendix-section-number}

\begin{figure*}[htb!]
    \centering
    \includegraphics[width=0.99\linewidth]{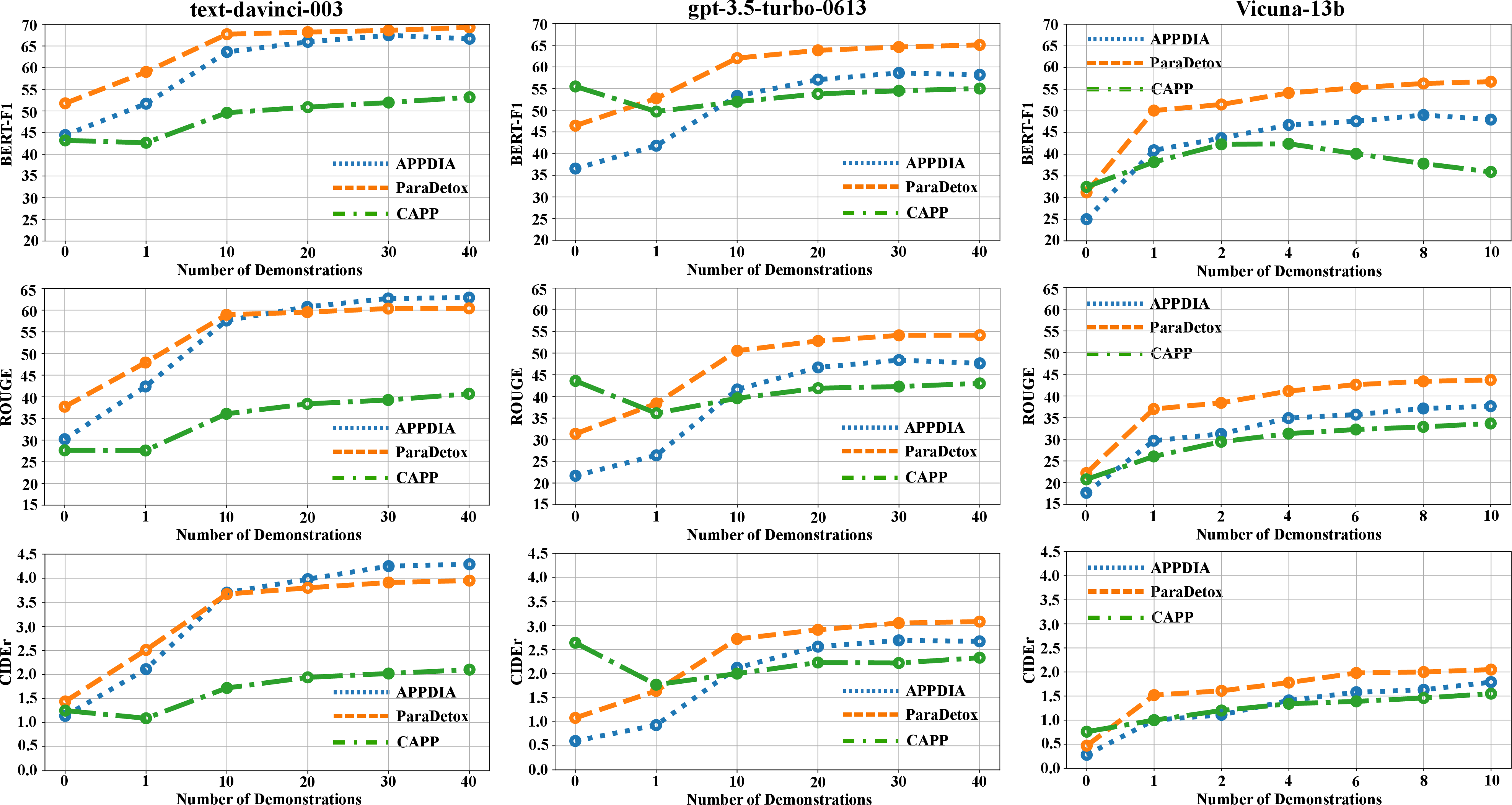}
    \caption{Performance using different evaluation metrics as a function of number of demonstrations used in the prompt. Noticeable improvement in score performance is observed in the beginning, with performance saturating after a certain number of demos.}
    \label{fig_appendix_number-of-examples}
\end{figure*}

Figure \ref{fig_appendix_number-of-examples} provides additional details, supporting the information described in Section \ref{subsection-number}.

\begin{table*}[ht!]
\centering
\scalebox{0.8}
{\begin{tabular}{|c|c|c|c|}
    \hline
    \textbf{Dataset} & \textbf{Method} & \makecell{\textbf{Similarity}\\ \textbf{Compute Time}$\downarrow$} & \makecell{\textbf{Demo Retrieval}\\ \textbf{Compute Time}$\downarrow$} \\
    \hline
    \hline
    \multirow{6}{*}{APPDIA} & MMR-BERT (10 Demos) &  4.7538  & 0.0792 \\
    \cline{2-4} & MMR-BERT (40 Demos) & 4.7538 & 0.5668  \\
    \cline{2-4}  & MMR-Embedding (10 Demos) & 0.0025 & 0.0786  \\
    \cline{2-4}  & MMR-Embedding (40 Demos) &  0.0025 & 0.5569 \\

    \cline{2-4}&  gpt-3.5-turbo-1106 (10 Demos) & \textbf{0.0025} & \textbf{0.0005}  \\
    \cline{2-4}&  gpt-3.5-turbo-1106 (40 Demos) & \textbf{0.0025} & \textbf{0.0005}   \\
    \hline
    \hline
    \multirow{6}{*}{ParaDetox} & MMR-BERT (10 Demos) &  4.7538 & 0.5990  \\
    \cline{2-4} & MMR-BERT (40 Demos)  & 4.7538  & 4.39 \\
    \cline{2-4}  & MMR-Embedding (10 Demos) & 0.0025 & 0.6010 \\
    \cline{2-4}  & MMR-Embedding (40 Demos) & 0.0025 & 4.17  \\

    \cline{2-4}&  gpt-3.5-turbo-1106 (10 Demos) & \textbf{0.0025} & \textbf{0.0049}  \\
    \cline{2-4}&  gpt-3.5-turbo-1106 (40 Demos) & \textbf{0.0025} & \textbf{0.0047}  \\
    \hline
    \hline
    \multirow{6}{*}{CAPP} & MMR-BERT (10 Demos)  &  4.7538  & 0.3937 \\
    \cline{2-4} & MMR-BERT (40 Demos) &  4.7538 & 2.97  \\
    \cline{2-4}  & MMR-Embedding (10 Demos) & 0.0025  & 0.3936 \\
    \cline{2-4}  & MMR-Embedding (40 Demos) & 0.0025  & 3.06 \\

    \cline{2-4}&  gpt-3.5-turbo-1106 (10 Demos) & \textbf{0.0025}   & \textbf{0.0031} \\
    \cline{2-4}&  gpt-3.5-turbo-1106 (40 Demos) & \textbf{0.0025}  & \textbf{0.0031}  \\
    \hline
    \end{tabular}}
    \caption{Comparison of the proposed  demonstration selection and ordering approach to MMR \cite{ye2022complementary}. Here, the proposed approach refers to the \emph{Most Similar (Descending Order)} approach outlined in Section \ref{subsection-order}. While MMR provides marginal performance gains on all three datasets as shown in Table \ref{table-comparison}, it is several orders of magnitude slower than the proposed approach. Here, the mean \emph{Similarity Compute Time} measures the average time taken to perform similarity measurement between a query test sample and all the available reference training samples; The mean \emph{Demo Retrieval Compute Time} measures the average time taken to select $n$ demonstrations from the available training set based on the similarity measurements done previously.}\label{table-select-order-comparison}
    \vspace{-0.0in}
\end{table*}

\subsection{Selection and Order of Demonstrations}\label{appendix-section-order}

This section provides additional details, supporting the information described in Section \ref{subsection-order}. While the original paper \cite{ye2022complementary} suggests using BERT-Score for similarity measurement, we also explore the cosine similarity measurement for the normalized embeddings extracted using the sentence transformer models. We refer to the original implementation as MMR-BERT and name the new variant as MMR-Embedding. The parameter $\lambda$ was set to 0.5 for both MMR approaches. Tables \ref{table-comparison} and \ref{table-select-order-comparison} compare the performance differences between the different MMR approaches and the proposed best performing ICL-based approach, \emph{i.e.}, the \emph{Most Similar (Descending Order)} method. Comparison is done using various quantitative evaluation metrics and compute time metrics. Note, for the proposed approach, the compute times reported correspond to the \emph{Most Similar (Descending Order)} method, however, these would also be the same for the other demonstration selection and ordering approaches described in Section \ref{subsection-order}. The different compute time metrics explored are defined as follows -- The mean \emph{Similarity Compute Time} measures the average time taken to perform similarity measurement between a query test sample and all the available reference training samples; The mean \emph{Demo Retrieval Compute Time} measures the average time taken to select $n$ demonstrations from the available training set based on the similarity measurements done previously. Here, $n$ is defined beside each method's name within parenthesis. The total time taken to process each query test sample would approximately be equal to the sum of the above two compute times. 

Table \ref{table-comparison} shows that the MMR methods offer marginal improvement with respect to the different quantitative metrics but at the expense of significant time delays. Note, the Similarity Compute Time for MMR-Embedding is the same as our proposed ICL-based approach since both employ the cosine similarity metric. Quantitative metrics for the two MMR approaches is only reported for 10 demonstrations in Table \ref{table-comparison} as we observed even greater delays with respect to Demo Retrieval Compute Time when trying to retrieve 40 demonstrations. Table \ref{table-select-order-comparison} reports the two Compute Times for the different MMR approaches and proposed ICL method, for both 10 and 40 demonstrations. Note, gpt-3.5-turbo-1106 was used as the generation model for MMR. For the proposed ICL-based approach we observe negligible differences in compute times when we want to retrieve either 10 or 40 demonstrations. However, the MMR-BERT approach becomes 10 times slower when trying to retrieve 40 demonstrations instead of 10.

While there are several demonstration selection and ordering approaches \cite{ye2022complementary,zhang2022active,lu2021fantastically} that can help push the performance ceiling, one must also make sure if these approaches can be easily implemented and scaled up in a real-time, real-world application setting. For example, \cite{zhang2022active} propose a reinforcement learning algorithm to identify generalizable policies to select demonstrations but find that this approach offers diminishing returns on larger, more sophisticated LLMs. The approach described by \cite{lu2021fantastically} to overcome few-shot prompt order sensitivity is better suited for multi-class classification tasks, since the entropy-based statistics framework discussed to identify performant prompts is not directly applicable to text generation tasks like paraphrasing.

\subsection{Significance of Instruction}\label{appendix-section-instruction-absence}

\begin{figure}[tb!]
    \centering
    \includegraphics[width=0.99\linewidth]{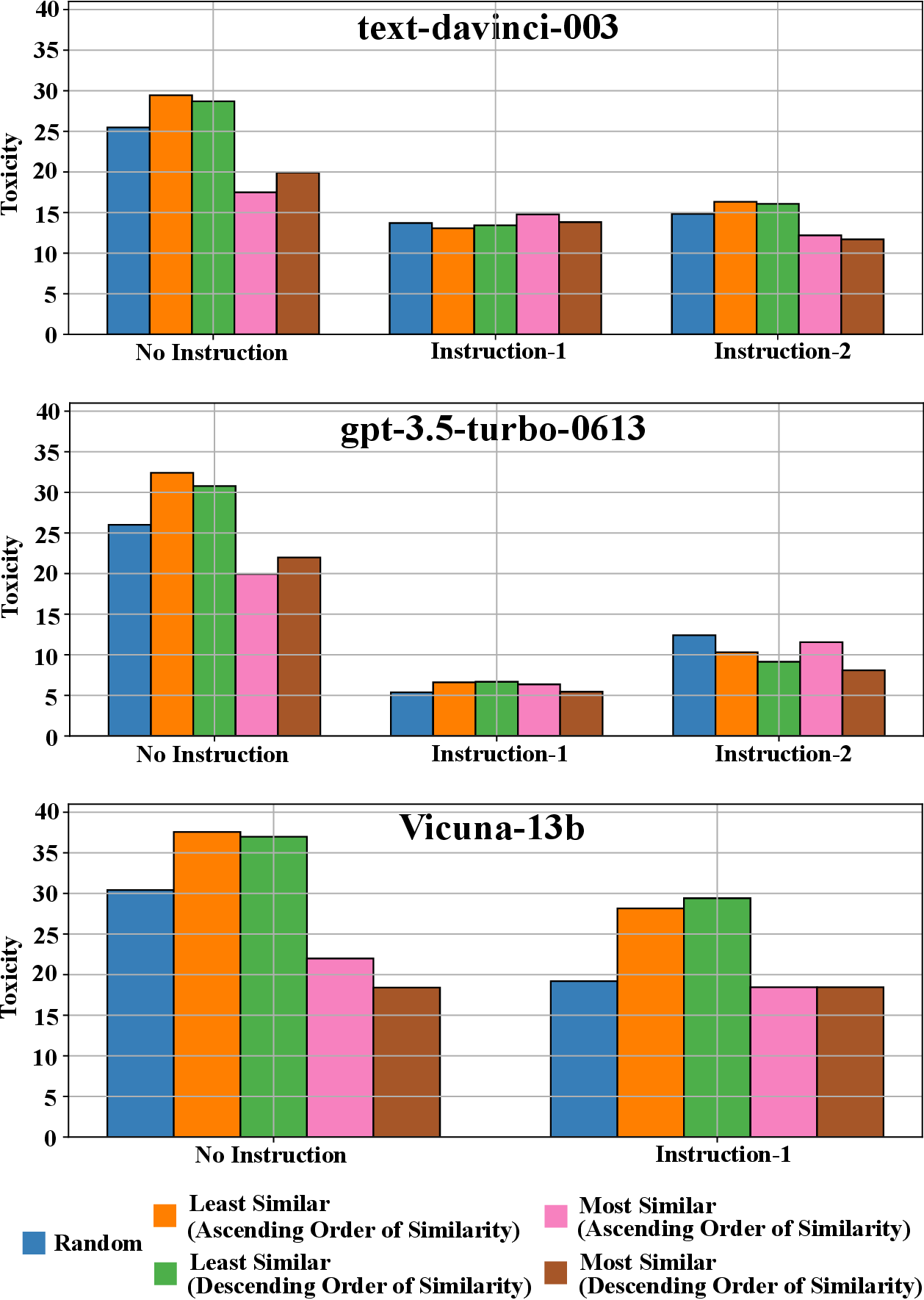}
    \caption{Measured toxicity performance of different models on the APPDIA dataset, with different instructions but with the same set of demos. No instruction setting results in paraphrases with higher toxicity.}
    \label{fig_appendix_bad_instruction}
\end{figure}

Figure \ref{fig_appendix_bad_instruction} provides additional details, supporting the information described in Section \ref{subsection-instruction-absence}.

\subsection{Robustness to Reduced Training Data}\label{appendix-subsection-reduced-training-data}

\begin{figure}[htb!]
    \centering
    \includegraphics[width=0.99\linewidth]{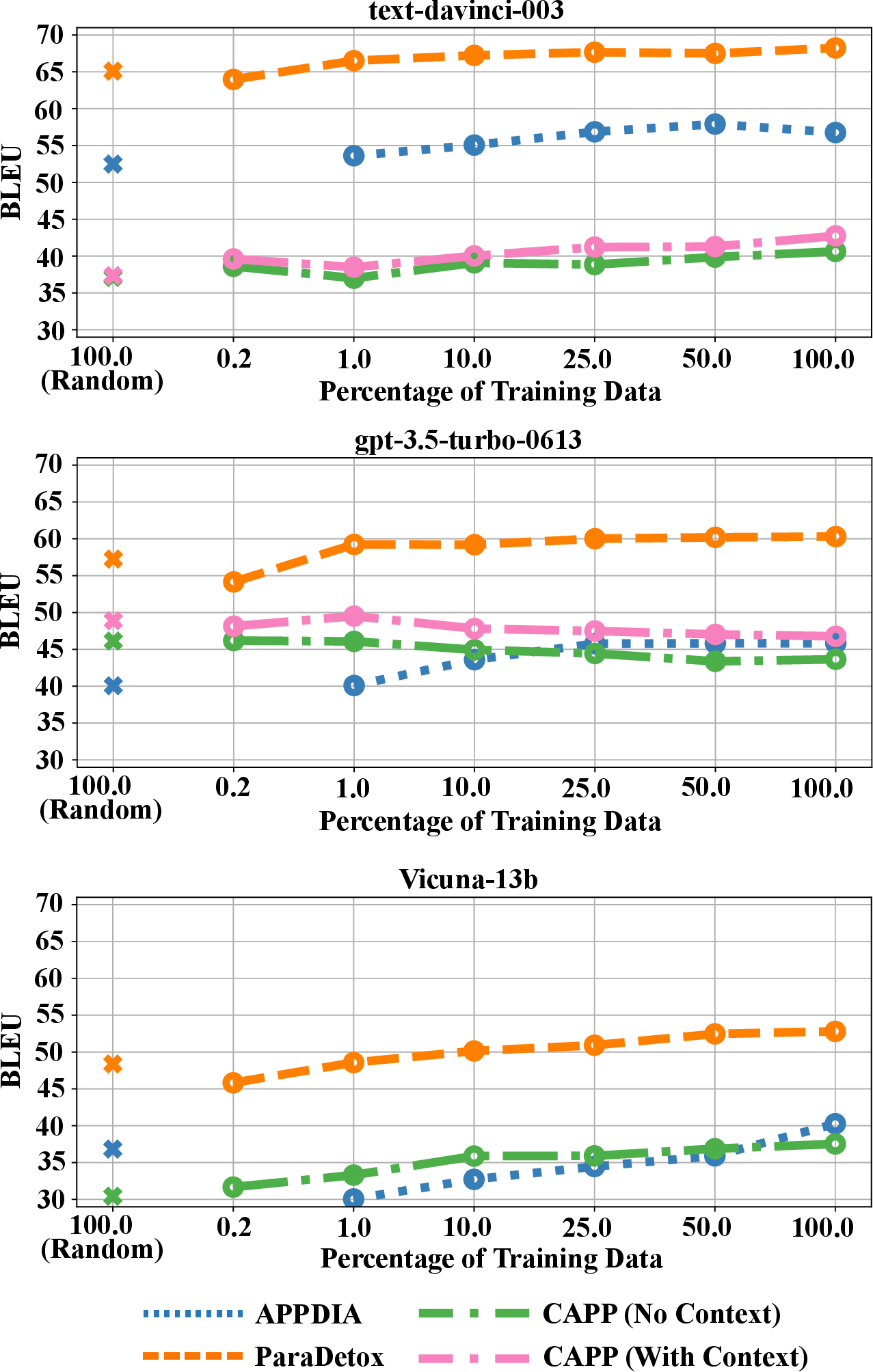}
    \caption{BLEU for the \emph{``Most Similar (Descending Order)''} approach as a function of percentage of training data available and comparison to Random demo selection with access to 100\% of the training data.}
    \label{fig_appendix_training-percentage}
\end{figure}

Figure \ref{fig_appendix_training-percentage} provides additional details, supporting the information described in Section \ref{subsection-reduced-training-data}.

\subsection{Correlation between Manual Evaluation Score and Automated Toxicity Metric}\label{appendix-subsection-correlation}

The manual quality evaluation score considers not just Toxicity minimization in the generated paraphrase but also considers meaning preservation. Directly comparing it to the automated toxicity score is not possible since there is a semantic mismatch between the two metrics. Instead, we first compute the difference in the toxicity measured between the (offensive) utterance and the (inoffensive) paraphrase. Next, we compute the Pearson correlation between this difference in toxicity score to the manual quality evaluation score. This difference captures the comparisons made by the annotator while coming up with the manual scoring as shown in Tables \ref{table-scoring_dataset} and \ref{table-scoring_paraphrase}. Therefore, if there is correlation between this automatically computed difference and the manual scores, then convergent validity is assured. 

Tables \ref{pearson-correlation-1}, \ref{pearson-correlation-2}, \ref{pearson-correlation-3} show the computed Pearson correlation coefficient between the manual quality evaluation score and automated toxicity score under two different settings. Type-1 represents the Pearson correlation coefficient between the manual quality score and automated toxicity score of the paraphrased output; and Type-2 represents the Pearson correlation coefficient between the manual quality score and difference in measured toxicity between original utterance and paraphrased output. Note, the dynamic range of the toxicity captured in the CAPP dataset is low because it mostly consists of rude speech that contains little or no foul language.

\begin{table*}[ht!]
\centering
\begin{minipage}[t]{.32\linewidth}
\centering
\scalebox{0.65}{
\begin{tabular}{|c||c||c|c|}
    \hline
    \textbf{Dataset} & \textbf{Method} & \textbf{Type-1} & \textbf{Type-2} \\
    \hline
    \hline
    \parbox[t]{2mm}{\multirow{15}{*}{\rotatebox[origin=c]{90}{APPDIA}}} & Gold Standard & -0.28 & 0.07 \\
    \cline{2-4}  & BART & -0.41 & 0.36 \\
    \cline{2-4}  & DialoGPT & -0.40 & 0.28 \\
    \cline{2-4}  & \thead{text-davinci-003\\(10 demos)} & -0.45 & 0.25 \\
    \cline{2-4}  & \thead{text-davinci-003\\(40 demos)} & -0.43 & 0.27 \\
    \cline{2-4}  & \thead{gpt-3.5-turbo-0613\\(10 demos)} & -0.39 & 0.26 \\
    \cline{2-4}  & \thead{gpt-3.5-turbo-0613\\(40 demos)} & -0.35 & 0.21 \\
    \cline{2-4}  & \thead{Vicuna-13b\\(4 demos)} & -0.42 & 0.24 \\
    \cline{2-4}  & \thead{Vicuna-13b\\(10 demos)} & -0.41 & 0.17 \\
    \hline
\end{tabular}
}
\caption{Computed Pearson correlation coefficient on the APPDIA dataset between manual quality score and automated toxicity score of paraphrased output, denoted as Type-1; manual quality score and difference in measured toxicity between original utterance and paraphrased output, denoted as Type-2.}
\label{pearson-correlation-1}
\end{minipage}%
\hfill
\begin{minipage}[t]{.32\linewidth}
\centering
\scalebox{0.65}{
\begin{tabular}{|c||c||c|c|}
    \hline
    \textbf{Dataset} & \textbf{Method} & \textbf{Type-1} & \textbf{Type-2} \\
    \hline
    \hline
    \parbox[t]{2mm}{\multirow{15}{*}{\rotatebox[origin=c]{90}{ParaDetox}}} & Gold Standard & -0.15 & 0.19 \\
    \cline{2-4}  & \multirow{2}{*}{BART} & \multirow{2}{*}{-0.60} & \multirow{2}{*}{0.56} \\
    & & & \\
    \cline{2-4}  & \thead{text-davinci-003\\(10 demos)} & -0.22 & 0.20 \\
    \cline{2-4}  & \thead{text-davinci-003\\(40 demos)} & -0.22 & 0.28 \\
    \cline{2-4}  & \thead{gpt-3.5-turbo-0613\\(10 demos)} & -0.24 & 0.25 \\
    \cline{2-4}  & \thead{gpt-3.5-turbo-0613\\(40 demos)} & -0.26 & 0.24 \\
    \cline{2-4}  & \thead{Vicuna-13b\\(4 demos)} & -0.24 & 0.26 \\
    \cline{2-4}  & \thead{Vicuna-13b\\(10 demos)} & -0.49 & 0.41 \\
    \hline
\end{tabular}
}
\caption{Computed Pearson correlation coefficient on the ParaDetox dataset between manual quality score and automated toxicity score of paraphrased output, denoted as Type-1; manual quality score and difference in measured toxicity between original utterance and paraphrased output, denoted as Type-2.}
\label{pearson-correlation-2}
\end{minipage}%
\hfill
\begin{minipage}[t]{.32\linewidth}
\centering
\scalebox{0.65}{
\begin{tabular}{|c||c||c|c|}
    \hline
    \textbf{Dataset} & \textbf{Method} & \textbf{Type-1} & \textbf{Type-2} \\
    \hline
    \hline
    \parbox[t]{2mm}{\multirow{15}{*}{\rotatebox[origin=c]{90}{CAPP}}} & Gold Standard & -0.139 & -0.049 \\
    \cline{2-4}  & BART & -0.189 & -0.094 \\
    \cline{2-4}  & T5 & -0.188 & -0.005 \\
    \cline{2-4}  & \thead{text-davinci-003\\(10 demos)} & -0.155 & 0.022 \\
    \cline{2-4}  & \thead{text-davinci-003\\(40 demos)} & -0.218 & -0.016 \\
    \cline{2-4}  & \thead{gpt-3.5-turbo-0613\\(10 demos)} & -0.197 & -0.077 \\
    \cline{2-4}  & \thead{gpt-3.5-turbo-0613\\(40 demos)} & -0.265 & -0.129 \\
    \cline{2-4}  & \thead{Vicuna-13b\\(4 demos)} & -0.096 & -0.188 \\
    \cline{2-4}  & \thead{Vicuna-13b\\(10 demos)} & -0.176 & -0.058 \\
    \hline
\end{tabular}
}
\caption{Computed Pearson correlation coefficient on the CAPP dataset between manual quality score and automated toxicity score of paraphrased output, denoted as Type-1; manual quality score and difference in measured toxicity between original utterance and paraphrased output, denoted as Type-2.}
\label{pearson-correlation-3}
\end{minipage}
\end{table*}

\end{document}